\def\year{2020}\relax
\documentclass[letterpaper]{article} 
\usepackage{aaai20}  
\usepackage{times}  
\usepackage{helvet} 
\usepackage{courier}  
\usepackage[hyphens]{url}  
\usepackage{graphicx} 
\usepackage{graphicx}  
\usepackage{amsmath}
\usepackage{amssymb}
\newcommand{\R}{\mathbb{R}}
\DeclareMathOperator*{\argmax}{arg\,max}
\usepackage{natbib}

\title{A Response Retrieval Approach for Dialogue Using a Multi-Attentive Transformer}
\author{\textbf{Matteo Antonio Senese},\textsuperscript{\rm 1,}\textsuperscript{\rm 2}  \textbf{Alberto Benincasa},\textsuperscript{\rm 1,}\textsuperscript{\rm 2} 
\textbf{Barbara Caputo},\textsuperscript{\rm 2}
\textbf{Giuseppe Rizzo}\textsuperscript{\rm 1}\\
\textsuperscript{\rm 1}LINKS Foundation,
\textsuperscript{\rm 2}Politecnico di Torino \\
\{matteo.senese, alberto.benincasa, giuseppe.rizzo\}@linksfoundation.com,
barbara.caputo@polito.it}

\begin{document}

\maketitle

\begin{abstract}
This paper presents our work for the ninth edition of the Dialogue System Technology Challenge (DSTC9). Our solution addresses the track number four: Simulated Interactive MultiModal Conversations. The task consists in providing an algorithm able to simulate a shopping assistant that supports the user with his/her requests. We address the task of response retrieval, that is the task of retrieving the most appropriate agent response from a pool of response candidates. Our approach makes use of a neural architecture based on transformer with a multi-attentive structure that conditions the response of the agent on the request made by the user and on the product the user is referring to. Final experiments on the SIMMC Fashion Dataset show that our approach achieves the second best scores on all the retrieval metrics defined by the organizers. The source code is available at https://github.com/D2KLab/dstc9-SIMMC. 
\end{abstract}

\section{Introduction}
Dialogues have always fueled the interest of the whole AI community. Conversation is the way humans interact to exchange information and thoughts and represents the ideal paradigm we shall use to interface with an artificial agent. Building agents that can correctly understand and coherently carry on conversations has been recognized as a turning point for AI~\citep{turing1950computing}. Understanding a conversation requires non-trivial Natural Language Understanding (NLU) capabilities of content referral from both past interactions (conversational knowledge) and external information (world knowledge). Additionally, in order to create an effective communication channel, the agents involved in a conversation must express their thoughts in a comprehensible manner from a syntactic and semantic perspective.
Over the last years major works have shown that deep learning approaches can address chit-chat conversations with a self-supervised pre-training \citep{adiwardana2020towards}. These models are able to generate responses to natural language utterances maintaining a semantic coherence. Unfortunately they are not specifically designed to address goal-oriented domain-specific conversations that cover the majority of use cases nowadays. The Dialogue System Technology Challenge (DSTC) \citep{williams2014dialog} series are a series of challenges that aim to improve the state-of-the-art AI techniques we use to handle conversations. 
In recent years the AI community started to work on dialogue scenarios that require a common visual grounding between the parts~\citep{das2017visual,kottur-etal-2017-natural}. This additional source of complexity is pushed by many recent breakthroughs in the fields of vision~\citep{krizhevsky2017imagenet,ren2015faster,he2017mask} and language~\citep{mikolov2013efficient,vaswani2017attention,devlin-etal-2019-bert}. Following this direction, the SIMMC track~\citep{moon-etal-2020-situated} at DSTC9~\citep{gunasekara2020overview} proposes a challenge concerning a multimodal shopping assistant to empower the research towards the next generation of conversational assistants. The scenario proposed in the fashion dataset consists of a user that interacts with a shopping virtual assistant by sending pictures of clothing he/she is looking at together with textual requests (e.g., ``Is this dress available in red?''). The assistant is provided with the clothing catalog containing all information about all the products referred (e.g., the price, the sizes and the colors available for a certain product). In this work we describe an approach to address the response retrieval sub-task that consists in retrieving the most appropriate agent response from a pool of candidates responses conditioned on both the user request and the referred object. In Section~\ref{sec:task-descr}, we provide an overview of the task we address. In Section~\ref{sec:approach}, we illustrate the transformer-based neural architecture~\citep{vaswani2017attention} we used focusing on the multi-attentive structure of each decoder block composed of three attention layers: a self-attention layer for the generated tokens, an attention layer for the current user request, and an additional attention layer for the item referred by the current turn. In Section~\ref{sec:evaluation}, we show the results of the experiments performed on the fashion dataset. Lastly, Section~\ref{sec:conclusion} concludes the paper.

\section{Task Description}
\label{sec:task-descr}
In this work we address the problem of response retrieval to provide a good answer to a request made by the user during a conversation. The response retrieval task consists in defining a policy to retrieve the best possible answer to a user request from a pool of candidate responses. This task is nowadays considered to be a valid alternative to the full generation of a response, which is still difficult to do in a goal-oriented paradigm (i.e., leading the conversation towards a goal, different from chit-chat) and prone to generalization problems. The assumption of response retrieval is that an acceptable answer is already present in a set of all the possible answers the system can give.
Typical experimental setups to assess the quality of the produced algorithm consists in the creation of synthetic pools by inserting, in a random position, the true answer to a request and fill the pool with random answers to other random requests. The optimal algorithm should then be able to always find the best true answer among the set of responses in the pool (if ambiguities or errors are not present). In the following section we describe the model we implemented to perform response retrieval.

\section{Approach}
\label{sec:approach}
In this Section, we describe our approach to solve the response retrieval sub-task. The response retrieval sub-task consists in retrieving the most appropriate response from a pool of candidates. To retrieve the response we proceed to assign a score to each candidate in the pool. The generation of this score is constrained on the actual candidate, the last user request and the object referred in the request. Finally, we consider the candidate achieving the highest score as the most appropriate response ranked by our model. In the following sections, we illustrate the encoder and the multi-attentive decoder we implemented to address this task.

\subsection{Encoder}
The generation of a score for each candidate in the response pool is a multi-constraints problem. One constraint is the semantic complementarity that the agent response must have with the request made by the user. For instance, the response to a price request must contain information about a price. The second constraint is to ground the response on the common object shared between the user and the agent. For instance the user request can refer to a particular dress in the catalog she/he is looking at. This grounding results in enriching the response with information coming from the object attributes. Thus, a response of the agent to a request referring to a particular dress must contain not only a generic price but specifically the price of the referred dress.\\
To force the two constraints we proceed to properly encode the user request and the object attributes in a continuous space representation. For this purpose we use a transformer encoder~\citep{vaswani2017attention} with $N$ blocks where each block is composed of a multi-head self-attention layer followed by a feed-forward neural network. This encoder produces contextual embeddings that are vectors capturing the context in which the word is present. The choice of contextual embeddings is driven by experimental evidences reported in the literature of the last years showing their superior performance over the traditional ones~\citep{devlin-etal-2019-bert,liu2019roberta,radford2019language}.\\
We define two functions: $T_{enc}^u$ refers to the transformer encoding function where the superscript $u$ identifies the instance of the encoder used for the user utterance, $\Gamma$ is the tokenizer function. The user request $U = (w_1, ..., w_l)$ with length $l$ is first tokenized in $p$ tokens with a tokenizer function such that $\Gamma(U) = (t_1, ..., t_p)$ and then encoded as in Equation \ref{eq1}
\begin{equation}
    T_{enc}^u(\Gamma(U)) = (u_1, u_2, u_3, ..., u_p)
    \label{eq1}
\end{equation}
where $u_i \; \epsilon \; \R^{D_1}$ are contextual embeddings for each token in $U$ and $p$ is the number of tokens contained in $U$ (it depends from the tokenizer function) and $D_1$ is the dimensionality of each embedding.\\
We represent the object referred in the request as a series of attributes where each attribute is composed by an attribute name and one or more attribute values (see Equation~\ref{eq2}). 
\begin{equation}
    attr_i = (attr_{i, name}, \{attr_{i, val_j}\}_{j=1}^m)
    \label{eq2}
\end{equation}
In order to encode the entire information of an object, for each of its attributes, we create a string of name-values composed of the attribute name followed by colon ``:'' and comma separated values for that attributes $``attr_{i, name}: attr_{i, val_1}, attr_{i, val_2}, ..., attr_{i, val_m}"$ (e.g., ``colors: red, blue''). We then concatenated all the name-values strings and separate them with a period ``.''. We force this fixed pattern to make the model able to learn how to contextualize the attribute values within the attribute name that always come before closest colon on the left (e.g., contextualize words like ``l, xl, m'' with the attribute type ``available sizes''). From the output of our object encoder, we extract a series of contextual embeddings as in Equation \ref{eq3}
\begin{equation}
    T_{enc}^o(\Gamma(O)) = (o_1, o_2, o_3, ..., o_q)
    \label{eq3}
\end{equation}
where the superscript $o$ identifies the encoder instance used for the object encoding, $O = \{attr_i\}_{i=1}^n$, $q$ is the number of tokens contained in the object string representation, $u_i \; \epsilon \; \R^{D_2}$ are the contextual embeddings for the object attributes, and $D_2$ is the dimensionality of each embedding.\\

\subsection{Decoder}
The two encoded inputs are used to constrain the score for each candidate response inside the pool. The scores are generated using a transformer decoder designed with a multi-attentive structure. The multi-attentive block is composed of three attention layers. All the attention layers compute the attention in the same way using the query, key, value attention proposed by~\citep{vaswani2017attention}. The attention is then computed as in Equation~\ref{eq4}:
\begin{equation}
    Attention(Q, K, V) = softmax(\frac{QK^T}{\sqrt{d_k}})V
    \label{eq4}
\end{equation}
where $d_k$ is the size of K and Q, K and V are produced by projecting the attention input pair $(x_i, x_j)$ (belonging to the input sequence) onto three learnable matrices $W^Q$, $W^K$ and $W^V$ (see Equation \ref{eq5})
\begin{equation}
    \begin{aligned}
        Q &= x_i W^Q \\
        K &= x_j W^K \\
        V &= x_j W^V
        \label{eq5}
    \end{aligned}
\end{equation}
The first layer is the original multi-head attention layer~\citep{vaswani2017attention} that computes the attention among all the pairs of the already generated tokens. This layer is provided with a masking mechanism to avoid the transformer to see the future target tokens during training. The second attention layer is a multi-head utterance attention computing the attention scores between the output of the preceding layer and the user request encoding produced by the encoder. The transformer attention is computed having $x_i = y_i^{self}\;,\; x_j = u_j$, where $y_i^{self}$ represents the i-th output of the preceding self-attention attention layer. This layer forces the first constraint: semantic complementarity between the user request and the agent response. The third attention layer is a multi-head object attention layer that computes the attention between the output of the preceding layer and the contextual embeddings of the object's attributes. This third layer forces the second constraint: enriching the response with the semantic information from the object. The attention is computed using the input pair $x_i = y_i^{utt}\;,\; x_j = o_j$, where $y_i^{utt}$ represents the i-th output of the utterance attention layer. The final architecture also presents highway connections~\citep{srivastava2015training} and layer normalization~\cite{ba2016layer} after each attention layer. The output of the transformer decoder is a sequence of latent vectors as in Equation~\ref{eq6}
\begin{equation}
    T_{dec}(u_1, .., u_p, o_1, ..., o_q, w1, ..., w_c) = (s_1, ..., s_c)
    \label{eq6}
\end{equation}
where $w_i$ are the tokens in the candidate response, $c$ is the number of tokens in the candidate response, $s_i \; \epsilon \; \R^{D_3}$ is the latent vector produced for each candidate token, and $D_3$ is the dimensionality of each vector.\\
The final sequence of output vectors $(s_1, ..., s_c)$ is fed to a softmax function that outputs a probability distribution over a vocabulary of words. The probability distribution over the set of words composing the candidate response is used to compute the score assigned to each candidate in the pool as explained in the next section.

\begin{figure*}[t]
\centering
\includegraphics[width=0.8\textwidth]{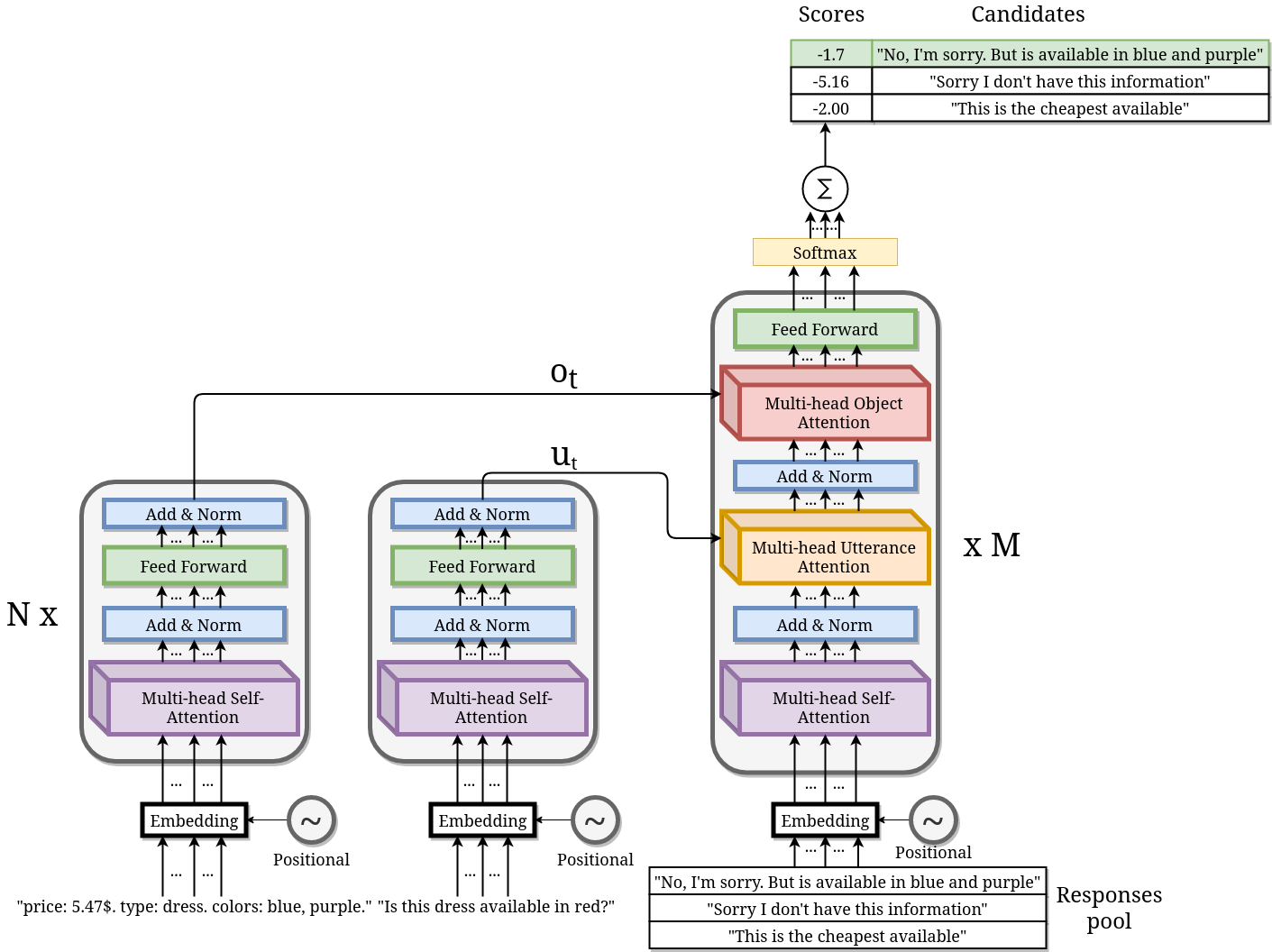} 
\caption{The final architecture consists in an encoder-decoder model. The two encoders encodes the last user request and the referred object attributes. The output of the two encoders $u_t$ and $o_t$ are two sequences of contextual embeddings. These embeddings are fed to the multi-attentive decoder together with the response candidates in the pool. The decoder outputs, one-by-one, the scores for each candidate using the normalized sum of the log-likelihoods of each word in the candidate. Finally the candidate with the highest score is considered as the true agent response retrieved by our model.}
\label{fig2}
\end{figure*}

\subsection{Scoring Criterion}
A  score for a candidate is computed considering the log-likelihood, estimated by the model, of each word in the candidate. The candidate response is extracted from the pool and fed as input to the decoder. The decoder produces the likelihood for each word, conditioned by the user request $U$, the referred object $O$ and the previous words in the candidate. For a candidate $X = (x_1, ..., x_c)$, the log-likelihood for the word $x_i$ is represented by Equation \ref{eq7}.
\begin{equation}
    LL(x_i) = log P(x_i | U, O, \{x_j\}_{j=1}^{i-1})
    \label{eq7}
\end{equation}
The final score for the entire candidate $X$ coincides with the log-likelihood to generate the candidate sequence. The full log-likelihood corresponds to the normalized sum of the log-likelihoods for each single word (see Equation \ref{eq8})
\begin{equation}
    SCORE_{LL}(X) = \sum_{j=1}^{c} \frac{LL(x_j)}{c}
    \label{eq8}
\end{equation}
where $c$ is the length of the candidate response $X$. The normalization factor is used to avoid a bias towards short candidates since each log-likelihood is a negative value and longer candidates have more negative contributions in the sum, thus resulting in smaller scores.\\
Additionally we add a grounding score to emphasize the grounding of the response within the referred object. In order to formalize this scoring function we define a set of attributes values for the object $O_{attrs} = \{attr_i^{val}\}_{i=1}^n$ and a function 
\begin{equation}
    \Omega(X, attr_i^{val})\;\epsilon\;\{1, 0\} 
\end{equation}
that takes as input the candidate response $X$ and an attribute value $attr_i^{val}$ and returns 1 if the response contains a reference to that attribute and 0 otherwise. In this work we choose $\Omega$ to be a regular expression. The grounding score is then defined as in Equation \ref{eq10} 
\begin{equation}
    SCORE_{gr}(X, O) = \sum_{j=1}^{n} \frac{\Omega(X, attr_j^{val})}{n}
    \label{eq10}
\end{equation}
The final score for a candidate $X$ is the sum of the log-likelihood score for each of the words in the candidate and the grounding score of the object attributes within the candidate (see Equation \ref{eq11}).
\begin{equation}
  SCORE(X, U, O) = SCORE_{LL}(X) + SCORE_{gr}(X, O)
  \label{eq11}
\end{equation}
The full architecture is shown in Figure \ref{fig2}.

\subsection{Optimization}
In order to train our model to compute log-likelihood of candidate responses we modeled the training stage as a text generation objective where the model is trained to generate the true agent response as in Equation \ref{eq12}
\begin{equation}
    \argmax_{\theta} P(w_t | w_1, ..., w_{t-1}, U, O; \theta)
    \label{eq12}
\end{equation}
where the $\{w_1, ..., w_t\}$ are words in the candidate response, U is the user request, O is the referred object and $\theta$ is the set model parameters. Each training step is a multi-class classification problem where the classes are the tokens in the output vocabulary. We force the model to minimize the cross-entropy loss for each word in the response candidate. Since minimizing the cross-entropy means minimizing the negative log-likelihood, using this type of loss allows us to maintain a compatibility of final objective between the training and the infer stage. We train our model for 3000 steps with a batch size of 256 and Adam optimizer~\citep{kingma2014adam} with learning rate of $10^{-3}$. We use the same encoder for request and object encoding $T_{enc}^u = T_{enc}^o$, with a number of layers $N=12$, model size $D_1=D_2=768$ and 12 attention heads. The encoder is initialized with pretrained BERT~\citep{devlin-etal-2019-bert} and its parameters are freezed during training. For the decoder we choose a number of layers $M=1$, model size $D_3=768$ and 4 heads for each attention layer. For $\Gamma()$ we choose a wordpiece tokenizer function~\citep{sennrich2015neural}.

\section{Evaluation}
\label{sec:evaluation}
In this section, we describe the SIMMC fashion dataset~\citep{moon-etal-2020-situated} we use for experiments and the results we obtained on the response retrieval task using our approach.

\subsection{Dataset}
The dataset we use for our experiments is the \textit{Fashion SIMMC Dataset}~\citep{moon-etal-2020-situated}. This dataset contains annotated dialogues collected in a Wizard-Of-Oz (WOZ) experiment between a user and a human that simulates a shopping assistant. A conversation starts with the user that sends a request to the assistant together with a picture of the clothing he/she is referring to. The conversation is multi-turn goal-oriented where the goal is to fulfill the user request supporting him/her throughout shopping. The dataset is provided with a catalog containing all the clothing objects together with their attributes. Each object is uniquely identified with an id and is enriched with attributes having a type and one or more values. Examples of attribute names are price, available size, color, type, brand. Values for attributes identify instances of the clothing object. For instance, available sizes can take values XL, XXL, S, XXS, thus indicating that the object is available in four different sizes. The dataset is provided in four splits: train, dev, test-dev and std-test. The train split is the one used for training and is composed of 3923 different dialogues and a total of 21176 turns. Dev split is intended to be used for hyperparameters tuning and is composed of 654 dialogues and 3.5k turns. The test-dev is the unofficial test split that contains annotations and is intended for internal testing. The test-std split is instead the official split (i.e., no annotations provided) used to compare the performance of the various submissions and to populate the leaderboard for DSTC9. In the next section we report the results on test-dev and test-std sets.

\begin{table}
    \begin{tabular}{|| c | c | c | c | c | c ||}
          \hline
          \textbf{} & \textbf{MRR} & \textbf{R@1} & \textbf{R@5} & \textbf{R@10} & \textbf{M.Rank} \\
          \hline
          baseline & $0.253$ & $16.3$ & $33.1$ & $41.7$ & $27.4$\\
          \hline
          team 1 & $0.346$ & $21.5$ & $48.8$ & $63.9$ & $15.1$\\
          \hline
          team 2 & $0.586$ & $46.2$ & $72.9$ & $83.2$ & $5.6$\\
          \hline
          team 3 & $0.379$ & $25.2$ & $50.5$ & $64.5$ & $14.3$\\
          \hline
          team 4 & \textbf{0.694} & \textbf{55.3} & \textbf{88.9} & \textbf{96.5} & \textbf{2.75}\\
          \hline
          \hline
          ours (w/o ss)& $0.398$ & $27.0$ & $54.0$ & $67.1$ & $14.1$\\
          \hline
          ours (w/ ss) & $0.419$ & $29.1$ & $55.9$ & $68.7$ & $13.6$\\ 
          \hline
     \end{tabular}
     \caption{Results on the test-dev split of \textit{SIMMC Fashion}.}
     \label{tab:dev-results}
\end{table}

\begin{table}
    \begin{tabular}{|| c | c | c | c | c | c ||} 
        \hline
        \textbf{} & \textbf{MRR} & \textbf{R@1} & \textbf{R@5} & \textbf{R@10} & \textbf{M.Rank} \\ [0.5ex] 
        \hline\hline
        baseline & $0.113$ & $4.5$ & $14.8$ & $22.1$ & $42.1$ \\ 
        \hline
        team 1 & $0.302$ & $18.1$ & $41.4$ & $56.1$ & $17$ \\
        \hline
        team 2 & $0.074$ & $2.2$ & $9.2$ & $14.3$ & $47.4$ \\
        \hline
        team 3 & $0.355$ & $22.8$ & $47.8$ & $62.6$ & $15.4$ \\
        \hline
        team 4 & \textbf{0.667} & \textbf{51.3} & \textbf{87.8} & \textbf{96.3} & \textbf{2.98} \\
        \hline
        \hline
        ours (w/ ss) & $0.39$ & $26.7$ & $52.1$ & $66$ & $14.8$ \\ 
        \hline
    \end{tabular}
    \caption{Results on the test-std split of \textit{SIMMC Fashion}.}
    \label{tab:std-results}
\end{table}

\subsection{Results}
The metrics proposed to assess the quality for response retrieval of the various submissions to SIMMC track are: the Mean Reciprocal Rank, the Mean Rank and the recalls at 1, 5 and 10. Tables \ref{tab:dev-results} and \ref{tab:std-results} report results on test-dev and test-std splits respectively. On test-std, that is the official test set for the SIMMC track leaderboard, we achieved the second best scores overall on all the 5 metrics. Our model outperforms the baselines with a gap of $+27.7\%$ on MRR, $+22.2\%$ on R@1, $+37.3\%$ on R@5,  $+43.9\%$ on R@10, and an improvement from $27.4$ to $13.6$ for the Mean Rank. In table \ref{tab:dev-results} we also report results for our model with and without the semantic score. The semantic score improves the results on the test-dev splits of $+2.1\%$ on MRR and R@1, $+1.9\%$ on R@5,  $+1.6\%$ on R@10, and $0.5$ points on the Mean Rank.

\section{Conclusion}
\label{sec:conclusion}
In this paper we have described our work for track number four of DSTC9 Challenge: Situated Interactive MultiModal Conversation (SIMMC). The aim of this track is to develop techniques to empower the next generation of situated multi-modal virtual assistants. We implemented a model that learns to estimate a score for the response retrieval task and we evaluated it on the Fashion dataset. The final model achieves the second best scores on all the metrics defined by the organizers on the Fashion dataset. In future experiments we will address the response retrieval task on similar conversational datasets using the same architecture to effectively assess the learning capabilities of transformers with a multi-attentive structure.

\bibliography{main}
\bibliographystyle{aaai}

\end{document}